\documentclass[letterpaper]{article}
\usepackage{aaai25}
\usepackage{times}
\usepackage{helvet}
\usepackage{graphicx}
\usepackage{tabularx}
\usepackage{booktabs}
\usepackage{multicol}
\usepackage{nameref}
\usepackage{natbib}
\usepackage{courier}
\usepackage{url}
\usepackage{xcolor}
\newcommand{\answerYes}[1]{\textcolor{blue}{#1}} 
\newcommand{\answerNo}[1]{\textcolor{teal}{#1}} 
\newcommand{\answerNA}[1]{\textcolor{gray}{#1}} 
 
\newcommand{\edge}[2]{#1 \rightarrow #2}

\renewcommand{\vec}[1]{\ensuremath \mathbf{#1}}
\newcommand{\embin}[0]{\ensuremath \vec{u}}
\newcommand{\basein}[0]{\ensuremath \vec{b}}
\newcommand{\resin}[0]{\ensuremath \vec{r}}

\newcommand{\embout}[0]{\ensuremath \vec{v}}

\newcommand{\lead}[0]{\ensuremath \textsc{Lead}}

\begin{document}
%
\title{Words and Action: \\
Modeling Linguistic Leadership in \#BlackLivesMatter Communities}
\author{
    Dani Roytburg\textsuperscript{\rm 1, \rm 2},
    Deborah Olorunisola\textsuperscript{\rm 3},
    Sandeep Soni\textsuperscript{\rm 1},
    Lauren Klein\textsuperscript{\rm 1, \rm 4} \\
}
\affiliations{
    \textsuperscript{\rm 1}Emory University, Department of Quantitative Theory \& Methods \\
    \textsuperscript{\rm 2}Emory University, Department of Computer Science \\
    \textsuperscript{\rm 3}Yale University, Department of Data Science and Statistics \\
    \textsuperscript{\rm 4}Emory University, Department of English \\
}
\maketitle
\begin{abstract}
In the wake of the 2024 US presidential election, pundits on both the left and the right pointed to a conservative backlash against ``woke politics'' to explain the election's outcome. These politics, rooted in substantive beliefs about equity and justice--and particularly racial justice--owe their most recent rise to prominence to the Black Lives Matter (BLM) movement. A significant body of work, both qualitative and quantitative, has documented how BLM was able to move these beliefs from the margin to the mainstream. In this paper, we focus on the words that index these beliefs, devising a novel method of modeling semantic leadership across a set of communities associated with the BLM movement that is informed by domain-specific theory about Black Twitter. We describe our bespoke approaches to time-binning, community clustering, and connecting communities over time, as well as our adaptation of state-of-the-art approaches to semantic change detection and semantic leadership induction. We find evidence at scale of the leadership role of BLM activists and progressives, as well as of Black celebrities. We also find evidence of sustained conservative engagement with this discourse, suggesting an alternative explanation for how we have arrived at the present political moment. 
\end{abstract}
\section{Introduction}
In early 2020, when the Black Lives Matter (BLM) movement had secured its place in the US national consciousness and when this project began, it was still necessary to provide evidence of how battles over social and political change are waged through both words and action. Writing again in April 2025, as we prepare this paper for final submission, it is evident to all (or should be) that words index larger concepts and debates. We need look no further than the US federal government, where ideas about equity and justice--and racial justice in particular--have not only become renewed topics of debate; the terms ``equity'' and ``racial justice'' themselves have been banned from government use \cite{yourish2025}. While this act of censorship serves as a negative example, ample evidence also points to how words can open up conversations, introduce new ideas into public consciousness, and positively impact social behavior and government policy alike \cite{dunivin_black_2022,Yan_Chiang_Lin_2024}. In this paper, we return to the Black Lives Matter movement, which we see as one of the most successful contemporary examples of the positive value of words (among other contributions), to explore with precision how words enter into a specific sociopolitical conversation, how they transform as that conversation evolves, and who is responsible for carrying forward those new or changed meanings.
\begin{figure}[h]
    \centering
    \includegraphics[width=0.95\linewidth]{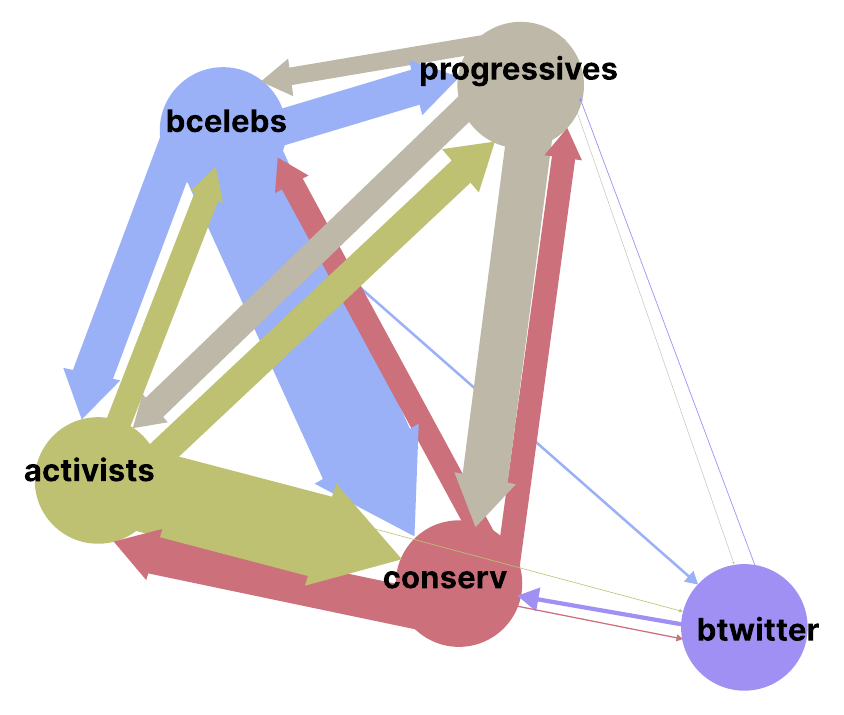}
    \caption{\textbf{Semantic leadership in \#BlackLivesMatter}: Nodes represent the most central communities of the BLM network. The thickness of the edges are proportional to the number of word changes shared by each pair.}
    \label{fig:overall_network}
\end{figure}

While there exists abundant research on Black Lives Matter (BLM), both as a hashtag~\cite{choudhury_social_2016,jackson2020,lekhac2022,giorgi_twitter_2022,Jones_Nurse_Li_2022} and a movement~\cite{freelon_beyond_2016,nytimesOluwatoyinSalau,clark2024}, there are still open questions about how precisely BLM was able to ``push the mainstream public sphere on issues of social progress'' \cite[p. xxvii]{jackson2020}. This is largely due to the unique community structure of Twitter, in which communities are defined not by individual users but by hashtags, and by the conversations among users that hashtags enable \cite{tufekci2013,spiro2016,jackson2020,brock2020}. Operationalizing these medium-specific theories about community formation on Twitter, as well as about ``Twitter time,'' we present a multipart model that allows us to 1) detect the latent communities associated with the \#BlackLivesMatter hashtag and link them over time; and 2) detect semantic changes associated with each community using word embeddings. Together with a measure of semantic leadership \cite{soni2021abolitionist}, we are able to induce a semantic leadership network (see Figure 1), allowing us to identify the communities responsible for introducing new or changed word meanings into the BLM network and the communities responsible for adopting (or co-opting) them.

We find that the community of BLM activists plays an outsized role in introducing semantic changes into the network, and that these changes are most consistently taken up by the conservative community. We find a similar albeit weaker signal between the progressive and conservative communities, confirming the role of left-leaning discourse in shaping the general terms of debate \cite{lekhac2022}. We also find that Black celebrities significantly shape the discourse, introducing word changes that the center/left news media as well as the conservative community later take up. The scale of our data, combined with our methods of validation, provide a new layer of evidence to support the largely qualitative scholarship that affirms the role of the BLM movement for bringing ideas about racial justice from the margin to the mainstream. 
Our results also provide an alternative explanation for how we have arrived at the present political moment. Contrary to a narrative of conservative backlash against a movement that became ``too woke'', we find that the conservative community --- as the largest follower of word changes across the entire timespan of our study --- had been engaging directly with the BLM movement from its very start. Today, we see the end goal of this engagement: to distort and disrupt the messaging of the BLM movement, and ultimately, to weaponize the movement's words against its most closely held beliefs. 

To summarize, our contributions are as follows:
\begin{itemize}
    \item Evidence at scale of how BLM activists shaped the discourse around racial justice, pushing new ideas from the margin to the mainstream 
    \item Early examples of specific terms (e.g. \textit{theory}) that have since become central to government policy and debate    
    \item A precise network structure of the communities that comprise the \#BlackLivesMatter hashtag, derived from extensive hand-labeling and domain-specific research
    \item An example of how domain-specific theory--here, media studies scholarship on Black Twitter--can inform technical modeling approaches 
\end{itemize}

\section{Background and Prior Work}

\begin{figure*}[t]
    \centering
    \includegraphics[width=1\linewidth]{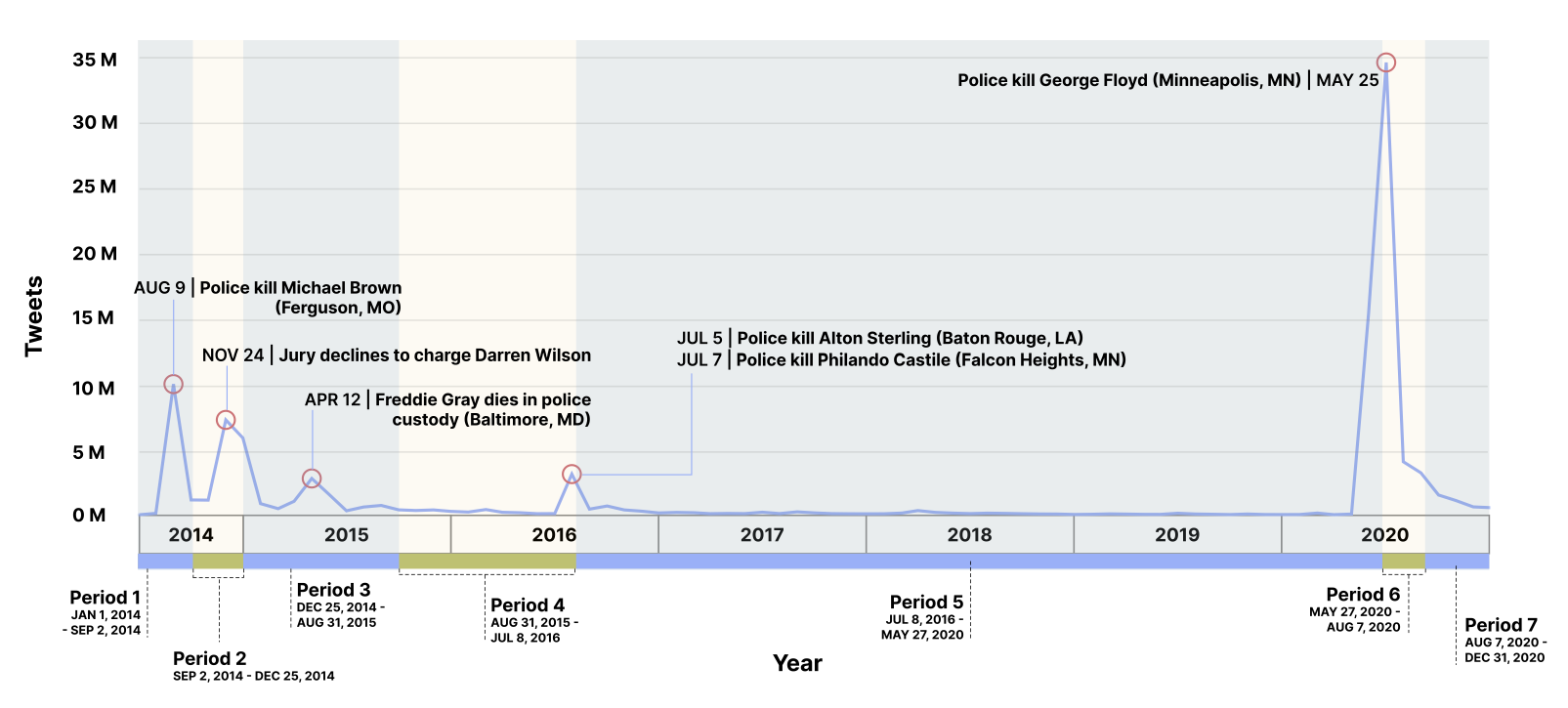}
    \caption{\textbf{Volume of tweets per month}: We visualize the temporal distribution of tweets in our dataset in the style of Freelon et al (\citeyear{freelon_beyond_2016}). The seven temporal partitions are indicated here both by color and a reference label (e.g., Period 1). Some pivotal events to the BLM movement in each partition are annotated.}
    \label{fig:timeline}
\end{figure*}

\subsection{Research on Black Twitter}
The topic of Twitter, and Black Twitter in particular, has been the subject of significant research in the fields of communication and media studies, among other humanities and social science fields. In \textit{Distributed Blackness} (\citeyear{brock2020}), André Brock theorizes the community structure of Black Twitter as consisting of the people who are online and in conversation with each other at any given time. He also theorizes ``Twitter time'' as non-linear, defined more by specific events and the bursts of conversation they prompt than by any standard timescale. In \textit{Hashtag Activism} (\citeyear{jackson2020}), Sarah J. Jackson, Moya Bailey, and Brooke Foucault Welles propose a similar conception of Twitter communities organized by the hashtag itself. Hashtags, they explain, ``designate collective thoughts, ideas, arguments, and experiences that might otherwise stand alone.'' Additional work on Black Twitter has considered the perspectives of the users themselves \cite{clark2024}, its historical antecedents \cite{klassen2021}, and its possible futures \cite{walcott2024}. This qualitative work constitutes the foundation for our project. Our contribution represents an attempt to formalize the concepts and structures that these media studies scholars describe. 

\subsection{Online Social Movements}
Several characteristics of online social movements, including the motivation and role of participants~\cite{tufekci2012social,Bozarth_Budak_2017,Jones_Nurse_Li_2022}, activist influence on potential recruits~\cite{gonzalez2011dynamics}, formation and sustenance of collective identities~\cite{brown_sayhername_2017}, and mobilization of activists~\cite{theocharis2015using,freelon_quantifying_2018,mundt_scaling_2018,Yan_Chiang_Lin_2024,Mendelsohn_Vijan_Card_Budak_2024}, have been extensively studied. BLM, in particular, has been studied as a unique example of a highly contested social movement with a sizeable online footprint~\cite{arif_acting_2018,stewart_drawing_2017,gallagher_divergent_2018,giorgi_twitter_2022}. The online discourse of this movement has been shaped by tragic offline events~\cite{peng_event-driven_2019} and has persisted for almost a decade~\cite{dunivin_black_2022}. In terms of our project, the closest related work is Freelon, McIlwain, and Clark's ``Beyond the Hashtags'' report (\citeyear{freelon_beyond_2016}), which employs community detection followed by hand-labeling by domain experts in order to trace the changes in network structure over the first 18 months of the BLM movement. Recognizing the knowledge and expertise involved in this project, our goal was to carry forward the authors' work. Here, we extend the timeframe of analysis by another five years; develop new methods that allow us to expand the initial set of communities labeled by their team; and introduce an additional layer of linguistic analysis that complements the original study.     

Linguistic methods have also been used to study different aspects of BLM. For example, the dynamics of participation have been shown to be correlated with textual features of tweets~\cite{choudhury_social_2016}; the impact of police killings has been shown to reflect in the emotional narratives of social media participants~\cite{field_analysis_2022}; hashtag use has been shown to draw attention~\cite{blevins_tweeting_2019,Yan_Chiang_Lin_2024} or push back against issues~\cite{gallagher_divergent_2018}. Framing and rhetorical strategies have been shown to be means of differentiation and coalition forming~\cite{stewart_drawing_2017,wilkins2019whose,Mendelsohn_Vijan_Card_Budak_2024}. We extend this line of work by inducing the leadership structure within the coalition of communities through the adoption of linguistic changes by community members.

\begin{table*}[t]
\centering
\small
\renewcommand{\arraystretch}{1.2} 
\begin{tabular}{llllll}
\toprule
Period & Time Span & Days & Pre-filtered Tweets & Post-filtered Tweets & Post-filtered Users\\
\midrule
1 & \texttt{Jan 01, 2014 - Sep 01, 2014} & 244 & 7,852,863 & 1,809,274 & 331,479 \\
2 & \texttt{Sep 02, 2014 - Dec 24, 2014} & 114 & 11,590,864 & 2,744,459 & 497,168 \\
3 & \texttt{Dec 25, 2014 - Aug 30, 2015} & 249 & 6,775,714 & 1,804,867 & 228,935 \\
4 & \texttt{Aug 31, 2015 - Jul 07, 2016} & 312 & 2,539,907 & 695,180 & 143,220 \\
5 & \texttt{Jul 08, 2016 - May 26, 2020} & 1419 & 7,196,284 & 1,630,376 & 347,567 \\
6 & \texttt{May 27, 2020 - Aug 06, 2020} & 72 & 41,593,336 & 5,252,117 & 1,659,510 \\
7 & \texttt{Aug 07, 2020 - Dec 31, 2020} & 146 & 5,112,396 & 877,086 & 214,050 \\
\bottomrule
\end{tabular}
\caption{\textbf{Descriptive statistics}: Tweet and user statistics for each temporal partition. Start and end dates are inclusive.}
\label{tab:twitter_bins}
\end{table*}

\subsection{Semantic Change and Leadership}
Sociocultural changes often get encoded as lexical semantic changes in language~\cite{tahmasebi2018survey}. Consequently, computational methods to automatically extract semantic changes, given a timestamped text collection, have been developed with increasing regularity~\cite[\textit{interalia}]{wijaya2011understanding,kim2014temporal,kulkarni2015statistically,hamilton2016diachronic,giulianelli2020analysing,card2023substitution}. Semantic changes detected by such methods are demonstrably effective in two ways: 1) as first-order objects of analytical interest, such as in tracking the evolution of political issues~\cite{rudolph2018dynamic} or shift in racial attitudes over time~\cite{garg2018word}; 2) as units that can be aggregated to recover the latent leadership structure among scholarly articles~\cite{soni2020follow,soni2022predicting} or newspapers~\cite{soni2021abolitionist}. In this work, we follow the latter approach, using word embeddings to unveil the leadership dynamics between communities and their role in shaping online discourse of the BLM movement. 

\section{Data}
Our dataset consists of tweets containing ``\#BlackLivesMatter'' between June 1, 2014 and December 31, 2020. These tweets were collected in two phases: 
tweets between June 1, 2014 and May 31, 2015 were rehydrated in April 2021 using \texttt{twarc}\footnote{https://twarc-project.readthedocs.io/en/latest/} from tweet IDs provided by Freelon et al. (~\citeyear{freelon_beyond_2016}).\footnote{In addition to the \#BlackLivesMatter hashtag, these tweets also included 44 additional ``keywords related to BLM and police killings of Black people under questionable circumstances'' \cite[p. 21]{freelon_beyond_2016}} Our choice of this data source was motivated by our desire to build upon the  hand-labeled communities in that paper, which we then used to seed our own community detection step (see next section). Tweets between June 1, 2015 and December 31, 2020, were collected using Twitter's search API between April and August 2021. Together, this resulted in a total of 82,661,364 tweets from 14,914,359 unique users. 

\subsection{Filtering}
Our analysis proceeds in two steps: a community detection step (see~\nameref{sec:community_detection}) and a semantic leadership network induction step (see~\nameref{sec:leadership_network_induction}). For the former, we filtered out all tweets tagged as non-English using \texttt{langid}~\cite{lui2012langid}; for the latter, we did additional filtering passes to remove tweets that were retweets, exact duplicates, or tagged as non-English by \texttt{langid}~\cite{lui2012langid} and \texttt{fasttext}~\cite{grave2018learning}. The first round of filtering left 69,560,572 tweets; the second round of filtering left 32,652,123 tweets. The breakdown of tweets by timebin and unique users is shown in Table~\ref{tab:twitter_bins}.

\subsection{Ethics and Privacy}
Throughout our analysis, we remain attentive to concerns about extracting knowledge and intellectual labor from online communities for academic research \cite{bruckman2002,jules2018,dym2020,jackson2020,walsh2023}. These concerns motivate our modeling approach, which is designed to extend rather than replace existing scholarship. In addition, our multiracial project team has members who have engaged with the BLM movement in various roles, on and offline. With respect to concerns about privacy and attribution, we are not required to identify individual users in most cases, as the majority of our analysis is undertaken at an aggregate level. In the four instances where we identify individual users or their tweets, we take a context-specific approach \cite{dym2020}. Figure \ref{fig:treemap} requires that we give a sense of the users that comprise each community. In this case, to avoid unwanted exposure, we rank users by in-degree and name only those within the top five of each community that a) have maintained public accounts through Twitter's transition to X; and b) as of September 2024, still meet the ``reasonably public'' threshold of 3000 or more followers as formulated by Freelon et al. (\citeyear{freelon_beyond_2016}). Table 2 requires that we provide examples of the word changes that our model detects. Here, we follow current best practice by paraphrasing tweets rather than quoting directly \cite{dym2020,mediumRecommendationsUsing} because we do not want to expose activist users or users with low follower counts to unwanted attention or possible harm \cite{jules2018}. We also confirm via the current X.com search interface that our ``ethically fabricated'' language \cite{markham2012} is not traceable back to the original user. The two users mentioned by name in the paper are public figures and as such, have upwards of one million followers.

\section{Community Detection and Matching}
\label{sec:community_detection}
Our first goal is to find relatively persistent communities of Twitter users throughout the full timespan of this study. In this section we describe the procedure for partitioning the timespan, mapping nodes to clusters within temporal partitions, and matching clusters across temporal partitions to find cohesive communities.

\subsection{Defining Discrete Time Periods}
We partition the dataset around the local maxima of tweet frequency, allowing irregular intervals. 
We find that the relative maxima correspond to real-world events relating to BLM. For instance, the first local maximum is found on August 11th, 2014, two days after the murder of Michael Brown ~\cite{pew_anderson,freelon_beyond_2016}. We use these real-world events as a qualitative baseline for validating possible maxima. When additional maxima occur within 30 days of another maximum, they are grouped into one period (see Table~\ref{tab:maxima}). From the 11 relative maxima that were extracted using this method, six became markers for time periods resulting in seven intervals (see Figure~\ref{fig:timeline} and Table~\ref{tab:twitter_bins}). 

\subsection{Time-Specific User Clustering}
\paragraph{User graphs} We use Freelon’s Twitter Subgraph Manipulator (TSM)\footnote{https://github.com/dfreelon/TSM} package to construct graphs from three forms of user-to-user interaction on Twitter found in tweets: replies, mentions, and retweets. These graphs are used to partition users into disjoint clusters in each of the seven temporal bins. 

\paragraph{Clustering objective} To obtain user clusters, we maximize intra-group modularity~\cite{newman_2004_finding,newman_2006_modularity} by modifying the Louvain clustering algorithm~\cite{blondel_fast_2008} --- a well-known algorithm for detection of an unconstrained number of communities --- using the TSM package. 

Because of our goal of extending the communities identified in Freelon, McIlwain, and Clark (\citeyear{freelon_beyond_2016}), which were hand-labeled by experts on BLM and Black Twitter, we augment the basic Louvain approach with six seed communities, consisting of 60 users, which were reconstructed from the communities documented in the original study. This improved the overall coherence of the clusters, which initial experiments had demonstrated to be poor.  

\paragraph{Clusters} We extract the top 50 clusters for each temporal bin, which can be seen as \textit{slices} of the larger BLM network, in the sense that they capture the time-bound representation of the BLM community structure.

\begin{figure*}[h]
    \begin{multicols}{2}
    \centering
    \includegraphics[width=0.985\linewidth]{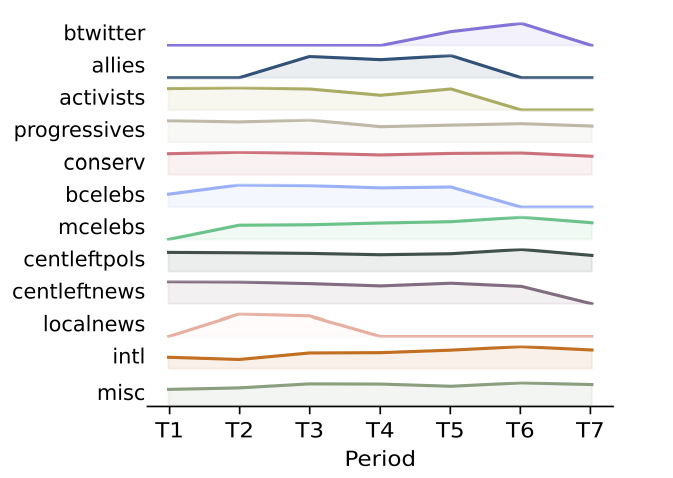}
    \caption{\textbf{Tweet counts over time} after filtering, by community. Grouping subcommunities allows for more consistent representation of each community over the entire span.}
    \label{fig:rivers}
    \hfill
\includegraphics[width=0.9\linewidth]{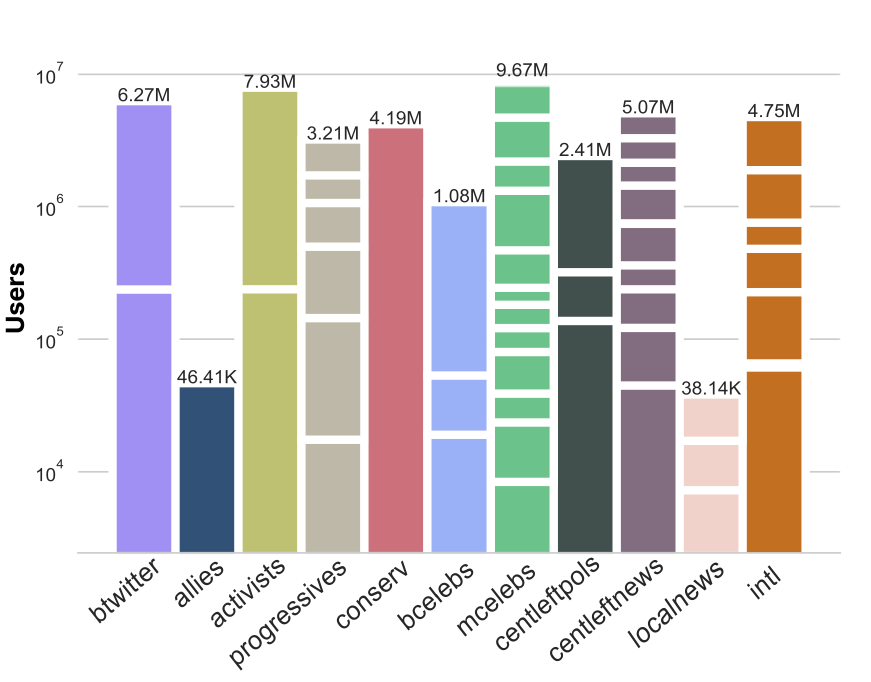}
\caption{\textbf{Community size}: Count of unique users in each community. Bar segments represent subcommunity clusters with height proportional to cluster size.} 
\label{fig:pack}
\end{multicols}
\end{figure*}
\subsection{From Clusters to Communities}
To find persistent communities from the ephemeral clusters obtained in the previous step, we cast the problem as a \textit{matching} problem in a bipartite graph. Specifically, for a pair of successive intervals, we construct a bipartite graph linking each cluster from one interval to every cluster in the next interval. The weight on these links represents similarity between the two clusters, and is calculated as a weighted Jaccard score taking into account the in-degree of the users in a cluster~\cite{freelon_2014_syria}. Following Freelon et al., we experimented with pruning the ``weaker'' edges if their similarity was below 0.3. However, since our intervals span over months or years, we see a greater level of fragmentation at this threshold. Consequently, we used a knee plot, which highlights the tradeoff between the number of matched communities and the jaccard threshold. Using this diagnostic, we lower the threshold to 0.07, a brightline that permits persistent communities with noise confined to spurious matches (see Figure~\ref{fig:knee}).

We repeat the matching process between each successive temporal interval, and throw out any intermediary edges with similarity less than 0.07, effectively identifying the chains of clusters that remain most persistent over time. In summary, the sets of disjoint, independent clusters are matched to the preceding and succeeding periods producing 65 cohesive and persistent communities (see Figure \ref{fig:rivers}).
\section{Labeling}
\subsection{Labeling Users}
With the sequenced slices matched, we can begin to develop qualitative characteristics for the 65 communities that result. We sought to generate a basic profile for each by surveying the top 20 users ranked by in-degree in each community, and giving each label from a set of iteratively redefined labels. The decision to use the top 20 users was heuristic; on average, the top 20 users made up 46.7\% of the total in-degree, demonstrating their centrality to each of the 65 communities. A sub-label is included which qualifies the sort of membership a user has. For instance, a CNN journalist like \textbf{@donlemon} would be classified as “Established Media”, sub-labeled “Journalist.” Each user is given a second optional label if the first is insufficiently representative; these labels are not hierarchical, as they are jointly important. From this set of classifications and from background research, we agree upon a set of categories which could maximally encompass our communities. 

In order to establish label agreement, the annotators constructed a series of pilot tests. 
These pilot test annotations yielded 83\% agreement between annotators on average.
With this established, each of the four annotators labeled 360 users. We randomly selected six communities (120 users) to be annotated by two annotators for agreement. Thus, annotators labeled 1,200 total users once and 120 twice to check for agreement.
We computed Fleiss's Kappa from the random pairwise assignment. Since some users received two non-ordinal labels, we present the maximum agreement score as reference. On average, interrater agreement was 0.798. After coming to consensus on how to define the mismatched labels, we continued from this user-level analysis to consolidating labels for the communities more broadly.

\subsection{Describing and Grouping Communities}
To develop a stronger sense of each community as a whole, the annotators used the set of labels assigned to its top 20 users to define a short description of each community. Using the two to four alternative descriptions for each community, the annotators then agreed on a final short description. 
In order to conduct semantic change analysis and perform permutation tests, it is necessary to condense the number of communities analyzed. Using the final short descriptions, we merged our 65 communities into 12 larger groups (see Figure ~\ref{fig:pack} and Figure ~\ref{fig:treemap}; ``miscellaneous'' not shown). 

We excluded two groups from this analysis, one for not being written in English and the other for containing spam. Since communities on Twitter change in membership over time, it is important to check if these group labels are apt. Once grouped into the 12 macro communities, the average in-degree of the top 20 users was no less than 20\%, suggesting that, relative to the size of their respective communities, these users play a central role in its dialogue. Thus, their presence provides important information about the background of other users and potential insight into how they may approach discussing social issues.

\section{Leadership Network Induction}
\label{sec:leadership_network_induction}
We adopted the method from~\citeauthor{soni2021abolitionist}~\shortcite{soni2021abolitionist} to induce a semantic leadership network between document sources. Soni et al. \citeyear{soni2021abolitionist} considered a set of historical newspapers as document sources and individual articles as documents. In contrast, we consider our Twitter user communities as document sources and individual tweets as documents.

Besides this superficial difference in the setup, there are two substantive differences. First, Soni et al. \citeyear{soni2021abolitionist} divided their timestamped collection of newspaper articles into temporal slices of equal timespans. In contrast, our tweet collection is divided into intervals demarcated by timestamps of real-world incidents. Second, Soni et al. \citeyear{soni2021abolitionist} considered document sources to remain fixed over time. In contrast, we consider communities of Twitter users as document sources which may undergo changes as a result of users joining, leaving and migrating between communities. We thus apply the methodology to our data but note that the interpretation of influence across communities should be qualified to account for these user dynamics.

For completeness, we briefly describe the core methodology here. Consider any tweet $i$ as a sequence of tokens (words) $w_i = (w_{i,1}, w_{i,2}, \ldots, w_{i, n_i})$ from a finite vocabulary $\mathcal{V}$, where $n_i$ indicates the tweet length. Our corpus consists of $N$ tweets, $\mathcal{W} = \{w_1, w_2, \ldots, w_N\}$. Each tweet has two labels: a discrete timestamp $t_i$, obtained by binning the document timestamps into one of $T$ bins; and a community label $s_i$ based on the community of the tweet's author.

\subsection{Semantic Change Detection}
To identify semantic changes, we learn temporal word-type embeddings from data. Other methods to detect semantic changes using token embeddings have been proposed but with mixed success~\cite{laicher2020cl}. Consequently, we approach the task as learning word embeddings and then enhancing them to account for information from temporal or other facets. Words are mapped to embeddings, typically by maximizing the following probability under the skipgram language model~\cite{mikolov2013distributed} between any pair of nearby tokens $(w_j, w_{j'})$ in a single document ,
\begin{equation}
P(w_{j'} \mid w_j) \propto \exp \left( \embout_{w_{j'}} \cdot \embin_{w_j} \right),
\label{eq:skipgram-prob}
\end{equation}
where $\embout_{w_{j'}}$ is the ``output'' embedding of $w_{j'}$, and $\embin_{w_j}$ is the ``input'' embedding of $w_{j}$; both set of embeddings are the parameters of the skipgram model that are learned from data. If documents are timestamped, then every input embedding can be split into a core embedding and a time-specific deviation as follows,
\begin{equation}
\label{eq:temporal-residual}
\embin_{w_{i,j}}^{(t_i)} = \basein_{w_{i,j}} + \resin_{w_{i,j}}^{(t_i)},
\end{equation}
where $\basein_{w_{i,j}}$ is the base embedding of the word $w_{i,j}$ and $\resin_{w_{i,j}}^{(t_i)}$ is the residual for time $t_i$. The residual turns the core meaning to a temporally specific meaning and are hence regularized towards zero to follow the assumption that for most words the core meaning remains intact over time. Embedding decomposition of this type has been used in other applications such as to identify geographic variation~\cite{bamman2014distributed} or perceptual differences in meaning~\cite{gillani2019simple}.

Once the base and the residual components of the input embeddings for each word are obtained, semantic changes can be found by comparing near-neighbors in the embedding space~\cite{hamilton2016cultural}. The upshot, at the end of this step, is a ranked order list of triples, consisting of the word that has changed, the timestamp for the onset of the change, and the timestamp for the conclusion of the change.   

\subsection{Identifying Semantic Leaders}
For each semantic change, we then score the communities that lead or lag that change. To do this, the input embeddings in Equation~\ref{eq:temporal-residual2} are modified to include a residual term for the community of each token. The input embedding is rewritten,
\begin{equation}
\label{eq:temporal-residual2}
\embin_{w_{i,j}}^{(t_i,s_i)} = \basein_{w_{i,j}} + \resin_{w_{i,j}}^{(t_i)} + \resin_{w_{i,j}}^{(t_i,s_i)},
\end{equation}
where $\resin_{w_{i,j}}^{(t_i,s_i)}$ is the source-specific temporal deviation added to the temporal and atemporal components of the input embedding. 

Next, a leadership score is calculated between a pair of communities  $s_1$ and $s_2$ and for a given change $(w, t_1, t_2)$, where $t_1 < t_2$ are the timestamps of a change in the meaning of word $w$, as a ratio of cross-correlation between the sources to the auto-correlation in meaning as follows,
\begin{equation}
\label{eq:lead-equation}
    \lead(\edge{s_1}{s_2},w,t_1,t_2) = \frac{\embin_{w}^{(t_1,s_1)} \cdot \embin_{w}^{(t_2, s_2)}}{\embin_{w}^{(t_1,s_2)} \cdot \embin_{w}^{(t_2,s_2)}}
\end{equation}

The tuple $(\edge{s_1}{s_2}, w, t_1, t_2)$ can be referred as a leadership event denoted by $e$. A higher $\lead(e)$ score indicates more leadership; a score of $1$ corresponds to a baseline case of no leadership. 

\subsection{Inducing Leadership Network}
\label{sec:method-control-randomness}
We can aggregate the leadership scores for each word obtained from Equation~\ref{eq:lead-equation} to construct a dense network between the communities. However, such a network may include changes that are, at best, spurious correlations due to random noise or structural biases such as temporal precedence of certain communities over others. To account for spurious correlations, we create $K=100$ randomized datasets by randomly swapping word tokens between communities. In principle, randomization breaks the link between individual communities and their contextual word statistics. Consequently, lead-lag relationships retained in the randomized dataset must be attributed to either structural bias or random noise. We filter out such spurious lead-lag relationships by comparing the lead score for a leadership event in the original non-randomized dataset against a set of lead scores for the same event in each of the randomized datasets. Any event in the final set of leadership events for further analysis satisfies the following criterion,
\begin{equation}
\lead(e) > \Phi_{.95}\left(\{\lead^{(k)}(e)\}_{k=1}^K \right),
\end{equation}
where the function $\Phi_{.95}(S)$ selects the 95th percentile value of the set $S$. The final output of the entire procedure is a list of events that can be aggregated to produce a weighted directed network between the communities, with the weight on any edge indicating the number of linguistic changes for which the pair of communities are in a lead-lag relationship of statistical significance. 

\section{Results}

\begin{table*}
\caption{\textbf{Semantic changes}: Examples of semantic changes with the tweets in which they appear, paraphrased from the original.}
\centering
\small
\renewcommand{\arraystretch}{1.2} 
\begin{tabular}{@{}lp{0.06\linewidth}p{0.40\linewidth}p{0.39\linewidth}}
\toprule
Word & Span & Earlier Tweet  & Later Tweet \\
\midrule
\textit{theory} & T3$\rightarrow$T7 & \textit{\#MikeBrown smearing follows classic a conspiracy \textbf{theory} pattern...} &  \textit{Critical Race \textbf{Theory} is an actual theory not just three words...} \\
\textit{active} & T2$\rightarrow$T6 & \textit{Downtown is \textbf{active} right now!} &  \textit{I'm calling for everyone, regardless of race, to get up and get \textbf{active}, and NOT BE SILENT!} \\
\textit{abolish} & T4$\rightarrow$T6 & \textit{Miscarriages of justice occur, especially to African Americans. This is precisely why the US should \textbf{abolish} executions!} &  \textit{\textbf{Abolish} the police. Their actions have shown for DECADES that they do not value Black lives.} \\
\textit{allyship} & T2$\rightarrow$T6 & \textit{\#AllLivesMatter is not \textbf{allyship}. It's not \*all\* lives that are in danger...} & \textit{A great convo on \textbf{allyship}, the importance of unlearning, and the dangers of white apathy...} \\
\textit{learning} & T1$\rightarrow$T6 & \textit{Just \textbf{learning} the details about \#MikeBrown. Heartsick for him, his family, his community, and all of society} & \textit{Listening. \textbf{Learning}. Acting. Here is an update on the steps we are taking} \\
\bottomrule
\end{tabular}
\label{tab:semantic_changes}
\end{table*}

Our approach offers insight into the changing meaning of conceptual keywords \cite{williams1995} related to political activism and the BLM movement, as well as about the specific communities that helped to spread those changed meanings. Here we discuss several sets of semantic changes identified by our model and the significance of the overall network we construct. Taken together, these results affirm the validity of our model and its underlying community clustering approach, as well as the crucial work done by BLM activists to shape conversations about police violence, antiblack racism, and possible solutions to both. They also affirm the importance of Black celebrities in communicating the ideas of BLM to the general public, as well as the presence of conservative critics who push back against activist discourse --- and, at times, co-opt its language. Below we discuss the findings from each phase of our methodological pipeline, building towards a conclusion about the decade-long lead-up to the current political climate in which the words of social justice and antiracism themselves have become the subject of legislative attention.

\subsection{Semantic Changes}
The first part of our model identifies the words that demonstrated statistically-significant semantic changes (Table \ref{tab:semantic_changes}). Beginning with general terms related to activism, we find that the term \textit{activists} undergoes a shift in meaning between 2014-2015, following the murders of Freddie Gray and Sandra Bland, when it has no particular connection to the work of BLM --- its near neighbors are terms like ``detained,'' ``targeted,'' and ``scuffle'' --- to the summer of 2020, by which point the term is very clearly associated with BLM activists in particular, as indicated by near neighbors including ``blackled,'' ``movements,'' ``bipoc,'' and ``advocates.'' The term \textit{action}, similarly, enters the dataset with no specific connotations of activism (its near neighbors are related to TV news channels) but, by the summer of 2020, has acquired specific valences of political change, as evidenced by the near neighbors of ``meaningful,'' ``tangible,'' ``steps,'' and ``stand.'' In a similar manner, the word \textit{active} also acquires connotations specific to allyship, as indicated by near neighbors of ``learning,'' ``listening,'' and ``unlearning,'' among others.  (The terms \textit{allyship} and \textit{learning} undergo similar changes, acquiring each other as near neighbors along similar timeframes). 

Terms like ``listening'' and ``learning'' point to more specific ways that BLM activists shaped the broader conversation about how to engage in activism around racial justice as well as how to contribute as an ally. But additional terms index specific issues and debates. For example, the term \textit{abolish} begins with associations with the death penalty and, in 2020, acquires an expanded meaning that encompasses the abolition of the police --- a key component of the BLM platform. The term \textit{abolishing}, similarly, acquires the near neighbors of ``defunding,'' ``reallocating,'' ``reforming,'' and ``demilitarization'' as the summer of 2020 approaches. We also find \textit{theory}, as in ``critical race theory,'' coalescing as a keyword during this time, acquiring associations with both ``crt'' and ``ideology.'' Taken together, these word changes point to a clear coalescing around the discourse of racial justice that we have come to associate with the movement. 

\begin{figure*}[h]
\centering
\includegraphics[width=.95\linewidth]{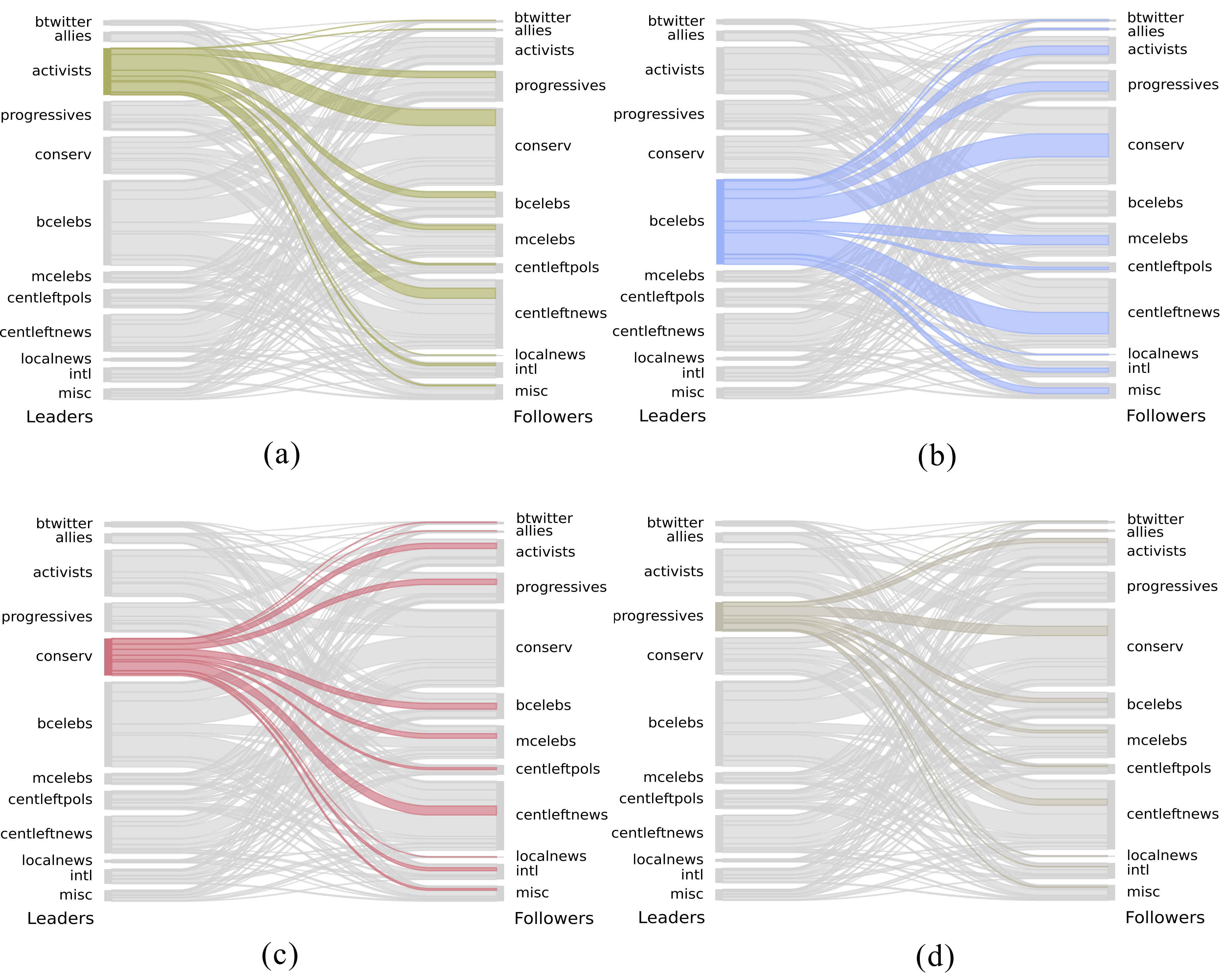}
    \caption{\textbf{Linguistic leadership}: The linguistic leadership of key communities including (a) activists, (b) Black celebrities, (c) conservatives, and (d) progressives with their contributions highlighted. The height of each bar is proportional to the overall number of words a community leads (left) or follows (right). The height of the connecting band is proportional to the number of words the community on the left leads the one on right.}
    \label{fig:common_caption}
\end{figure*}
\subsection{Leadership Network}
Unexpected insights begin to emerge when looking at the leadership network overall. To be sure, we observe the influence of the BLM activists (Figure 5a), who claim an outsized number of word changes in comparison to most other communities. The progressive community, which sits slightly closer to the center than the activist community, can count a significant number of word changes as well (Figure 5d), and it would make sense that the pattern of the progressives would track that of the BLM activists but less strongly. What is perhaps unexpected is that the conservative community --- moreso than the center/left news media --- is the most substantial follower of both the activist and progressive communities. While we might expect a more gradual path from the left to the center, and then to the right, our method shows how the conservative community engaged directly with the activist and progressive communities from the start. 

Another unexpected but not surprising result was the strong role of the community of Black celebrities in communicating the ideas of the BLM movement to the general public (Figure 5b). Indeed, Black celebrities claim 24.73\% of all of word changes, and are followed by activists, then center/left news media, and conservatives in high proportion. This finding suggests that the direct association with the BLM movement may mark its activists for pushback --- and, at times, outright aggression --- by the conservative community, the celebrity status of those in the Black celebrity community may protect them from some of the conservative vitriol, and as a result, allow them to communicate the ideas of the activists to a broader public. We see this in their leadership of terms like \textit{activism}, which transforms from the specific context of social movements to a word that itself indexes a role of social importance, as indicated by the near neighbors of ``importance,'' ``career,'' ``societal,'' and ``impact.'' We also see this in the terms that one would seem to associate with the BLM movement itself, such as \textit{listen}. The connotations of learning and unlearning, as well as amplifying the voices of those with direct experience-as indicated by the near neighbors of ``learn,'' ``unlearn,'' ``amplify,'' and ``explain,'' are communicated to the center/left news media commmunity from the Black celebrities. In addition, we find this community influencing the community of progressives, passing along the changed meaning of \textit{theory}, discussed above, as well as the more basic concept of \textit{equity}. Interestingly, the conservative community at times learns directly from the Black celebrity community. In particular, the term \textit{nonracist} is introduced by this community, and is adopted directly from the conservative community from there. 

A final observation has to do with the role of the conservative community in introducing new word meanings-and not just adopting them (Figure 5c). While few of these words are conceptual keywords, their overall number is substantial. This finding confirms the consistent and persistent engagement of these ideas by the conservative community --- not only in the wake of the summer of 2020, but from the movement's start. That these words are not conceptual keywords at once confirms prior findings about how conservatives --- as well as white people as a group --- demonstrate a lead-lag relationship with BLM activists \cite{lekhac2022} and shows how they engaged at the level of discourse: through the wide array of words that make up a language.  

\section{Implications and Limitations}
The wide range of statistically significant words in our dataset that are associated with political activism and with the BLM movement in particular, as well as the communities that they were introduced and/or adopted by, confirm the effectiveness of our approach for detecting semantic change and measuring leadership within a large and heterogeneous Twitter community. It also provides an additional layer of evidence for the valuable work of BLM activists, as well as Black celebrities, in advancing the cause of racial justice over the past decade. More recently, as Twitter has become X, and has prompted the scattering of the diverse communities who previously conversed together, we may need to consider additional adaptations to our method of detecting semantic change, just as we will need to reimagine the ways in which activists can spread their messages to a broader public and enlist others in their cause.

As far as the prominence of the conservative community in our study, we see significant implications for our understanding of the emergence of the ``anti-woke'' and ``anti-DEI'' agenda that, at the time of this writing, has entered legislation in nine US states \cite{brookings2021} and appears poised to wholly reshape government agencies and funding at the federal level. Whereas the common narrative is that this movement emerged in the aftermath of the summer of 2020, and has since been fueled by the Republican party, our findings show that there has always been direct conservative engagement with the discourse of Black Lives Matter. Moving forward, this finding may reshape the narrative of ``conservative backlash'' and, instead, point to a story of continual pushback and even outright aggression that met the movement from its start. 

Finally, we would like to highlight a key absence in this project, which is the missing data that might otherwise document the crucial role of the young Black activists who fueled the movement from its start. These regular young people were essential in the movement's early phases, as they were in organizing the on-the-ground protests that took place in every city where a police killing took place. Qualitative researchers have documented their crucial role \cite{jackson2020,clark2024}, as have news reports \cite{malalaBlackLivesMatterYoung}, yet their status as regular young people has meant that they lack the protections granted to celebrities or even to adults. As activists, they have been subjected to harassment and intimidation, and in some cases they have even been killed by the police themselves \cite{chicagotribunePuzzlingNumber,nytimesOluwatoyinSalau}. As early as mid-2015, when Freelon et al released their initial findings on the network structure of \#BLM, they found that the young Black activists who had been present in their data in 2014 had disappeared eighteen months later \cite{freelon_beyond_2016}. Using the usernames of their young Black activist community as seeds, we found similarly that these users had largely disappeared-save for the few who themselves became famous, and joined the Black celebrity community instead. 

In her influential theorization of \textit{missing data sets}, the artist and educator Mimi Onuoha explains how the gaps in datasets provide ``cultural and colloquial hints of what is deemed important,'' and further observes how ``this lack of data typically correlates with issues affecting those who are most vulnerable in that context'' \cite{githubGitHubMimiOnuohamissingdatasets}. In our \#BlackLivesMatter dataset and the findings that it has enabled, these young Black activists constitute this precise form of missing data. Thus as we celebrate our findings for how they affirm the role of the Black Lives Matter movement in shaping a larger public conversation, we must also hold space for what our study cannot confirm, which is the equally crucial role played by the activists whose words are missing from our dataset. Even if their words are unrecorded, they are not unremembered.

\section{Conclusion}
In this project we describe a method of modeling semantic leadership across a set of communities associated with the Black Lives Matter movement, which has been informed by qualitative research on the structure of social media and of Black Twitter in particular. We describe our bespoke approaches to time-binning, community clustering, and connecting communities over time, as well as our adaptation of state-of-the-art approaches to semantic change detection and semantic leadership induction. We find substantial evidence of the leadership role of BLM activists and progressives, as well as Black celebrities. 
We also find evidence of the sustained engagement of the conservative community with this discourse, suggesting an alternative explanation for how we have arrived at the present political moment, in which ``anti-woke'' and ``anti-DEI'' policy is being enacted nation-wide. 
Contrary to the dominant narrative of conservative backlash against a movement that became too radical for the center to support, we find that the conservative community had been engaging directly with the BLM movement from its very start. The conservative community is the largest follower of word changes across the entire timespan of our study, and introduces many word changes as well, albeit few original conceptual keywords. 

From our present vantage-point, in Spring 2025, we can now see the result of this sustained conservative engagement: a distortion of the meaning of the words that Black Lives Matter worked so hard to bring to public attention, ultimately weaponizing the most powerful of these words to lead to policies that dismantle, rather than more fully realize, the movement's original goals. With this in mind, just as we recognize the role of the young Black activists who are missing from our data, we also recognize and hold space for the valuable work of all those involved in the Black Lives Matter movement. While the immediate gains of their work are currently being dismantled, the ideas behind their words remain a powerful force. We submit this paper as evidence of what words can accomplish when coupled with a commitment to action and an unwavering belief in the importance of equality and justice for all.  

\section{ Acknowledgments}
Thank you to Tanvi Sharma for her design work on the final visualizations. This project would not have been possible without the data collection work of Alex Fan. It also greatly benefited from discussions with André Brock and Jacob Eisenstein. This project has been supported by a grant from the Mellon Foundation (G-2211-14240) and by Emory's Department of Quantitative Theory \& Methods.

\bibliography{aaai25}

\begin{thebibliography}{68}
\providecommand{\natexlab}[1]{#1}

\bibitem[{Anderson(2016)}]{pew_anderson}
Anderson, M. 2016.
\newblock The hashtag \#{BlackLivesMatter} emerges: {Social} activism on {Twitter}.
\newblock Technical report, Pew Research Center.

\bibitem[{Arif, Stewart, and Starbird(2018)}]{arif_acting_2018}
Arif, A.; Stewart, L.~G.; and Starbird, K. 2018.
\newblock Acting the {Part}: {Examining} {Information} {Operations} {Within} \#{BlackLivesMatter} {Discourse}.
\newblock In \emph{Proceedings of the ACM on Human-Computer Interaction}, CSCW, 1--27.

\bibitem[{Bamman, Dyer, and Smith(2014)}]{bamman2014distributed}
Bamman, D.; Dyer, C.; and Smith, N.~A. 2014.
\newblock Distributed representations of geographically situated language.
\newblock In \emph{Proceedings of the 52nd Annual Meeting of the Association for Computational Linguistics (Volume 2: Short Papers)}, 828--834.

\bibitem[{Blevins et~al.(2019)Blevins, Lee, McCabe, and Edgerton}]{blevins_tweeting_2019}
Blevins, J.~L.; Lee, J.~J.; McCabe, E.~E.; and Edgerton, E. 2019.
\newblock Tweeting for social justice in \#{Ferguson}: {Affective} discourse in {Twitter} hashtags.
\newblock \emph{New Media \& Society}, 21(7): 1636--1653.

\bibitem[{Blondel et~al.(2008)Blondel, Guillaume, Lambiotte, and Lefebvre}]{blondel_fast_2008}
Blondel, V.~D.; Guillaume, J.-L.; Lambiotte, R.; and Lefebvre, E. 2008.
\newblock Fast unfolding of communities in large networks.
\newblock \emph{Journal of Statistical Mechanics: Theory and Experiment}, 2008(10): P10008.

\bibitem[{Bozarth and Budak(2017)}]{Bozarth_Budak_2017}
Bozarth, L.; and Budak, C. 2017.
\newblock Is Slacktivism Underrated? Measuring the Value of Slacktivists for Online Social Movements.
\newblock \emph{Proceedings of the International AAAI Conference on Web and Social Media}, 11(1): 484--487.

\bibitem[{Brock(2020)}]{brock2020}
Brock, A.~J. 2020.
\newblock \emph{Distributed Blackness: African American Cybercultures}.
\newblock NYU Press.

\bibitem[{Brown et~al.(2017)Brown, Ray, Summers, and Fraistat}]{brown_sayhername_2017}
Brown, M.; Ray, R.; Summers, E.; and Fraistat, N. 2017.
\newblock \#{SayHerName}: {A} case study of intersectional social media activism.
\newblock \emph{Ethnic and Racial Studies}, 40(11): 1831--1846.

\bibitem[{Bruckman(2002)}]{bruckman2002}
Bruckman, A. 2002.
\newblock “{S}tudying the amateur artist: A perspective on disguising data collected in human subjects research on the Internet".
\newblock \emph{Ethics and Information Technology}, 4.

\bibitem[{Card(2023)}]{card2023substitution}
Card, D. 2023.
\newblock Substitution-based Semantic Change Detection using Contextual Embeddings.
\newblock In \emph{The 61st Annual Meeting of the Association for Computational Linguistics}.

\bibitem[{Choudhury et~al.(2016)Choudhury, Jhaver, Sugar, and Weber}]{choudhury_social_2016}
Choudhury, M.~D.; Jhaver, S.; Sugar, B.; and Weber, I. 2016.
\newblock Social {Media} {Participation} in an {Activist} {Movement} for {Racial} {Equality}.
\newblock In \emph{Proceedings of the International AAAI Conference on Web and Social Media}, volume~10, 92--101.

\bibitem[{Clark(2024)}]{clark2024}
Clark, M.~D. 2024.
\newblock \emph{"We Tried to Tell Y'All: Black Twitter and the Rise of Digital Counternarratives"}.
\newblock Oxford University Press.

\bibitem[{del Rio(2020)}]{nytimesOluwatoyinSalau}
del Rio, G. M.~N. 2020.
\newblock Oluwatoyin Salau, Missing Black Lives Matter Activist, Is Found Dead.
\newblock \url{https://www.nytimes.com/2020/06/15/us/oluwatoyin-salau-dead-aaron-glee.html}.
\newblock Accessed 2024-15-05.

\bibitem[{Dunivin et~al.(2022)Dunivin, Yan, Ince, and Rojas}]{dunivin_black_2022}
Dunivin, Z.~O.; Yan, H.~Y.; Ince, J.; and Rojas, F. 2022.
\newblock Black {Lives} {Matter} protests shift public discourse.
\newblock \emph{Proceedings of the National Academy of Sciences}, 119(10): e2117320119.

\bibitem[{{D}ym and Fiesler(2020)}]{dym2020}
{D}ym, B.; and Fiesler, C. 2020.
\newblock “{E}thical and privacy considerations for research using online fandom data".
\newblock \emph{Transformative Works and Cultures}, 33.

\bibitem[{Field et~al.(2022)Field, Park, Theophilo, Watson-Daniels, and Tsvetkov}]{field_analysis_2022}
Field, A.; Park, C.~Y.; Theophilo, A.; Watson-Daniels, J.; and Tsvetkov, Y. 2022.
\newblock An analysis of emotions and the prominence of positivity in \#{BlackLivesMatter} tweets.
\newblock \emph{Proceedings of the National Academy of Sciences}, 119(35): e2205767119.

\bibitem[{{FORCE11}(2020)}]{fair}
{FORCE11}. 2020.
\newblock The FAIR Data principles.
\newblock https://force11.org/info/the-fair-data-principles/.

\bibitem[{Freelon, Lynch, and Aday(2015)}]{freelon_2014_syria}
Freelon, D.; Lynch, M.; and Aday, S. 2015.
\newblock Online Fragmentation in Wartime: A Longitudinal Analysis of Tweets about Syria, 2011–2013.
\newblock \emph{The ANNALS of the American Academy of Political and Social Science}, 659(1): 166--179.

\bibitem[{Freelon, McIlwain, and Clark(2018)}]{freelon_quantifying_2018}
Freelon, D.; McIlwain, C.; and Clark, M. 2018.
\newblock Quantifying the power and consequences of social media protest.
\newblock \emph{New Media \& Society}, 20(3): 990--1011.

\bibitem[{Freelon, McIlwain, and Clark(2016)}]{freelon_beyond_2016}
Freelon, D.; McIlwain, C.~D.; and Clark, M. 2016.
\newblock Beyond the {Hashtags}: \#{Ferguson}, \#{Blacklivesmatter}, and the {Online} {Struggle} for {Offline} {Justice}.

\bibitem[{Gallagher et~al.(2018)Gallagher, Reagan, Danforth, and Dodds}]{gallagher_divergent_2018}
Gallagher, R.~J.; Reagan, A.~J.; Danforth, C.~M.; and Dodds, P.~S. 2018.
\newblock Divergent discourse between protests and counter-protests: \#{BlackLivesMatter} and \#{AllLivesMatter}.
\newblock \emph{PLOS ONE}, 13(4): e0195644.

\bibitem[{Garg et~al.(2018)Garg, Schiebinger, Jurafsky, and Zou}]{garg2018word}
Garg, N.; Schiebinger, L.; Jurafsky, D.; and Zou, J. 2018.
\newblock Word embeddings quantify 100 years of gender and ethnic stereotypes.
\newblock \emph{Proceedings of the National Academy of Sciences}, 115(16): E3635--E3644.

\bibitem[{Gebru et~al.(2021)Gebru, Morgenstern, Vecchione, Vaughan, Wallach, Iii, and Crawford}]{gebru2021datasheets}
Gebru, T.; Morgenstern, J.; Vecchione, B.; Vaughan, J.~W.; Wallach, H.; Iii, H.~D.; and Crawford, K. 2021.
\newblock Datasheets for datasets.
\newblock \emph{Communications of the ACM}, 64(12): 86--92.

\bibitem[{Gillani and Levy(2019)}]{gillani2019simple}
Gillani, N.; and Levy, R. 2019.
\newblock Simple dynamic word embeddings for mapping perceptions in the public sphere.
\newblock In \emph{Proceedings of the Third Workshop on Natural Language Processing and Computational Social Science}, 94--99.

\bibitem[{Giorgi et~al.(2022)Giorgi, Guntuku, Himelein-Wachowiak, Kwarteng, Hwang, Rahman, and Curtis}]{giorgi_twitter_2022}
Giorgi, S.; Guntuku, S.~C.; Himelein-Wachowiak, M.; Kwarteng, A.; Hwang, S.; Rahman, M.; and Curtis, B. 2022.
\newblock Twitter {Corpus} of the \#{BlackLivesMatter} {Movement} and {Counter} {Protests}: 2013 to 2021.
\newblock \emph{Proceedings of the International AAAI Conference on Web and Social Media}, 16: 1228--1235.

\bibitem[{Giulianelli, Del~Tredici, and Fern{\'a}ndez(2020)}]{giulianelli2020analysing}
Giulianelli, M.; Del~Tredici, M.; and Fern{\'a}ndez, R. 2020.
\newblock Analysing Lexical Semantic Change with Contextualised Word Representations.
\newblock In \emph{Proceedings of the 58th Annual Meeting of the Association for Computational Linguistics}, 3960--3973. Online: Association for Computational Linguistics.

\bibitem[{Gonz{\'a}lez-Bail{\'o}n et~al.(2011)Gonz{\'a}lez-Bail{\'o}n, Borge-Holthoefer, Rivero, and Moreno}]{gonzalez2011dynamics}
Gonz{\'a}lez-Bail{\'o}n, S.; Borge-Holthoefer, J.; Rivero, A.; and Moreno, Y. 2011.
\newblock The dynamics of protest recruitment through an online network.
\newblock \emph{Scientific reports}, 1(1): 1--7.

\bibitem[{Grave et~al.(2018)Grave, Bojanowski, Gupta, Joulin, and Mikolov}]{grave2018learning}
Grave, {\'E}.; Bojanowski, P.; Gupta, P.; Joulin, A.; and Mikolov, T. 2018.
\newblock Learning Word Vectors for 157 Languages.
\newblock In \emph{Proceedings of the Eleventh International Conference on Language Resources and Evaluation (LREC 2018)}.

\bibitem[{Hamilton, Leskovec, and Jurafsky(2016{\natexlab{a}})}]{hamilton2016cultural}
Hamilton, W.~L.; Leskovec, J.; and Jurafsky, D. 2016{\natexlab{a}}.
\newblock Cultural Shift or Linguistic Drift? Comparing Two Computational Measures of Semantic Change.
\newblock In \emph{Proceedings of the Conference on Empirical Methods in Natural Language Processing}, volume 2016, 2116.

\bibitem[{Hamilton, Leskovec, and Jurafsky(2016{\natexlab{b}})}]{hamilton2016diachronic}
Hamilton, W.~L.; Leskovec, J.; and Jurafsky, D. 2016{\natexlab{b}}.
\newblock Diachronic Word Embeddings Reveal Statistical Laws of Semantic Change.
\newblock In \emph{Proceedings of the 54th Annual Meeting of the Association for Computational Linguistics (Volume 1: Long Papers)}, volume~1, 1489--1501.

\bibitem[{Jackson, Bailey, and Welles(2020)}]{jackson2020}
Jackson, S.~J.; Bailey, M.; and Welles, B.~F. 2020.
\newblock \emph{HashtagActivism: Networks of Race and Gender Justice}.
\newblock MIT Press.

\bibitem[{Jones, Nurse, and Li(2022)}]{Jones_Nurse_Li_2022}
Jones, K.; Nurse, J.~R.; and Li, S. 2022.
\newblock Out of the Shadows: Analyzing Anonymous' Twitter Resurgence during the 2020 Black Lives Matter Protests.
\newblock \emph{Proceedings of the International AAAI Conference on Web and Social Media}, 16(1): 417--428.

\bibitem[{Jules, Summers, and Mitchell(2018)}]{jules2018}
Jules, B.; Summers, E.; and Mitchell, V. 2018.
\newblock Ethical Considerations for Archiving Social Media Content Generated by Contemporary Social Movements: Challenges, Opportunities, and Recommendations.
\newblock Technical report, Shift Collective.

\bibitem[{Kim et~al.(2014)Kim, Chiu, Hanaki, Hegde, and Petrov}]{kim2014temporal}
Kim, Y.; Chiu, Y.-I.; Hanaki, K.; Hegde, D.; and Petrov, S. 2014.
\newblock Temporal Analysis of Language through Neural Language Models.
\newblock In \emph{Proceedings of the ACL 2014 Workshop on Language Technologies and Computational Social Science}, 61--65.

\bibitem[{Klassen et~al.(2022)Klassen, Kingsley, McCall, Weinberg, and Fiesler}]{klassen2021}
Klassen, S.; Kingsley, S.; McCall, K.; Weinberg, J.; and Fiesler, C. 2022.
\newblock Black Lives, Green Books, and Blue Checks: Comparing the Content of the Negro Motorist Green Book to the Content on Black Twitter.
\newblock \emph{Proc. ACM Hum.-Comput. Interact.}, 6.

\bibitem[{Kulkarni et~al.(2015)Kulkarni, Al-Rfou, Perozzi, and Skiena}]{kulkarni2015statistically}
Kulkarni, V.; Al-Rfou, R.; Perozzi, B.; and Skiena, S. 2015.
\newblock Statistically significant detection of linguistic change.
\newblock In \emph{Proceedings of the 24th International Conference on World Wide Web}, 625--635.

\bibitem[{Laicher et~al.(2020)Laicher, Baldissin, Castañeda, Schlechtweg, and im~Walde}]{laicher2020cl}
Laicher, S.; Baldissin, G.; Castañeda, E.; Schlechtweg, D.; and im~Walde, S.~S. 2020.
\newblock CL-IMS @ DIACR-Ita: Volente o Nolente: BERT does not outperform SGNS on Semantic Change Detection.
\newblock arXiv:2011.07247.

\bibitem[{Le-Khac, Antoniak, and So(2023)}]{lekhac2022}
Le-Khac, L.; Antoniak, M.; and So, R.~J. 2023.
\newblock \#BLM Insurgent Discourse, White Structures of Feeling and the Fate of the 2020 "Racial Awakening".
\newblock \emph{New Literary History}, 53(4): 667--692.

\bibitem[{Lui and Baldwin(2012)}]{lui2012langid}
Lui, M.; and Baldwin, T. 2012.
\newblock langid. py: An off-the-shelf language identification tool.
\newblock In \emph{Proceedings of the ACL 2012 system demonstrations}, 25--30.

\bibitem[{Markham(2012)}]{markham2012}
Markham, A. 2012.
\newblock FABRICATION AS ETHICAL PRACTICE.
\newblock \emph{Information, Communication \& Society}, 15(3): 334--353.

\bibitem[{Mendelsohn et~al.(2024)Mendelsohn, Vijan, Card, and Budak}]{Mendelsohn_Vijan_Card_Budak_2024}
Mendelsohn, J.; Vijan, M.; Card, D.; and Budak, C. 2024.
\newblock Framing Social Movements on Social Media: Unpacking Diagnostic, Prognostic, and Motivational Strategies.
\newblock \emph{Journal of Quantitative Description: Digital Media}, 4.

\bibitem[{Mikolov et~al.(2013)Mikolov, Sutskever, Chen, Corrado, and Dean}]{mikolov2013distributed}
Mikolov, T.; Sutskever, I.; Chen, K.; Corrado, G.~S.; and Dean, J. 2013.
\newblock Distributed representations of words and phrases and their compositionality.
\newblock In \emph{Advances in neural information processing systems}, 3111--3119.

\bibitem[{Mundt, Ross, and Burnett(2018)}]{mundt_scaling_2018}
Mundt, M.; Ross, K.; and Burnett, C.~M. 2018.
\newblock Scaling {Social} {Movements} {Through} {Social} {Media}: {The} {Case} of {Black} {Lives} {Matter}.
\newblock \emph{Social Media + Society}, 4(4): 2056305118807911.

\bibitem[{Newman(2006)}]{newman_2006_modularity}
Newman, M.~E. 2006.
\newblock Modularity and community structure in networks.
\newblock \emph{Proceedings of the national academy of sciences}, 103(23): 8577--8582.

\bibitem[{Newman and Girvan(2004)}]{newman_2004_finding}
Newman, M.~E.; and Girvan, M. 2004.
\newblock Finding and evaluating community structure in networks.
\newblock \emph{Physical review E}, 69(2): 026113.

\bibitem[{Onuoha(2018)}]{githubGitHubMimiOnuohamissingdatasets}
Onuoha, M. 2018.
\newblock {G}it{H}ub - {M}imi{O}nuoha/missing-datasets: {A}n overview and exploration of the concept of missing datasets. --- github.com.
\newblock \url{https://github.com/MimiOnuoha/missing-datasets}.
\newblock Accessed: 2024-15-05.

\bibitem[{Peng, Budak, and Romero(2019)}]{peng_event-driven_2019}
Peng, H.; Budak, C.; and Romero, D.~M. 2019.
\newblock Event-{Driven} {Analysis} of {Crowd} {Dynamics} in the {Black} {Lives} {Matter} {Online} {Social} {Movement}.
\newblock In \emph{The {World} {Wide} {Web} {Conference}}, {WWW} '19, 3137--3143. New York, NY, USA: Association for Computing Machinery.
\newblock ISBN 978-1-4503-6674-8.

\bibitem[{Press(2019)}]{chicagotribunePuzzlingNumber}
Press, A. 2019.
\newblock A puzzling number of men tied to the Ferguson protests have since died.
\newblock \url{https://www.chicagotribune.com/2019/03/18/a-puzzling-number-of-men-tied-to-the-ferguson-protests-have-since-died/}.
\newblock Accessed: 2024-05-15.

\bibitem[{Ray and Gibbons(2021)}]{brookings2021}
Ray, R.; and Gibbons, A. 2021.
\newblock "Why are states banning critical race theory?".
\newblock \url{https://www.brookings.edu/articles/why-are-states-banning-critical-race-theory/}.
\newblock [Accessed 15-05-2024].

\bibitem[{Rudolph and Blei(2018)}]{rudolph2018dynamic}
Rudolph, M.; and Blei, D. 2018.
\newblock Dynamic embeddings for language evolution.
\newblock In \emph{Proceedings of the 2018 World Wide Web Conference}, 1003--1011.

\bibitem[{Soni, Bamman, and Eisenstein(2022)}]{soni2022predicting}
Soni, S.; Bamman, D.; and Eisenstein, J. 2022.
\newblock Predicting Long-Term Citations from Short-Term Linguistic Influence.
\newblock \emph{Findings of the Association for Computational Linguistics: EMNLP 2022}.

\bibitem[{Soni, Klein, and Eisenstein(2021)}]{soni2021abolitionist}
Soni, S.; Klein, L.; and Eisenstein, J. 2021.
\newblock Abolitionist Networks: Modeling Language Change in Nineteenth-Century Activist Newspapers.
\newblock \emph{Journal of Cultural Analytics}.

\bibitem[{Soni, Lerman, and Eisenstein(2020)}]{soni2020follow}
Soni, S.; Lerman, K.; and Eisenstein, J. 2020.
\newblock Follow the Leader: Documents on the Leading Edge of Semantic Change Get More Citations.
\newblock \emph{Journal for the Association of Information Science and Technology}.

\bibitem[{Spiro and Monroy-Hernández(2016)}]{spiro2016}
Spiro, E.; and Monroy-Hernández, A. 2016.
\newblock Shifting Stakes: Understanding the Dynamic Roles of Individuals and Organizations in Social Media Protests.
\newblock \emph{PLoS One}, 11(10).

\bibitem[{Stewart et~al.(2017)Stewart, Arif, Nied, Spiro, and Starbird}]{stewart_drawing_2017}
Stewart, L.~G.; Arif, A.; Nied, A.~C.; Spiro, E.~S.; and Starbird, K. 2017.
\newblock Drawing the {Lines} of {Contention}: {Networked} {Frame} {Contests} {Within} \#{BlackLivesMatter} {Discourse}.
\newblock \emph{Proceedings of the ACM on Human-Computer Interaction}, 1(CSCW): 96:1--96:23.

\bibitem[{Tahmasebi, Borin, and Jatowt(2019)}]{tahmasebi2018survey}
Tahmasebi, N.; Borin, L.; and Jatowt, A. 2019.
\newblock Survey of Computational Approaches to Lexical Semantic Change.
\newblock arXiv:1811.06278.

\bibitem[{Theocharis et~al.(2015)Theocharis, Lowe, Van~Deth, and Garc{\'\i}a-Albacete}]{theocharis2015using}
Theocharis, Y.; Lowe, W.; Van~Deth, J.~W.; and Garc{\'\i}a-Albacete, G. 2015.
\newblock Using Twitter to mobilize protest action: online mobilization patterns and action repertoires in the Occupy Wall Street, Indignados, and Aganaktismenoi movements.
\newblock \emph{Information, Communication \& Society}, 18(2): 202--220.

\bibitem[{Tufekci(2013)}]{tufekci2013}
Tufekci, Z. 2013.
\newblock ``{N}ot {T}his {O}ne'': Social Movements, the Attention Economy, and Microcelebrity Networked Activism.
\newblock \emph{American Behavioral Scientist}, 57(7): 848--870.

\bibitem[{Tufekci and Wilson(2012)}]{tufekci2012social}
Tufekci, Z.; and Wilson, C. 2012.
\newblock Social media and the decision to participate in political protest: Observations from Tahrir Square.
\newblock \emph{Journal of communication}, 62(2): 363--379.

\bibitem[{Walcott(2024)}]{walcott2024}
Walcott, R. 2024.
\newblock \#RIP Twitter: The Conditions of Black Social Media Platform Migration.
\newblock [Accessed 15-05-2024].

\bibitem[{Walsh(2023)}]{walsh2023}
Walsh, M. 2023.
\newblock \emph{Debates in the Digital Humanities 2023}, chapter The Challenges and Possibilities of Social Media Data: New Directions in Literary Studies and the Digital Humanities.
\newblock Minneapolis: University of Minnesota Press.

\bibitem[{Walsh(2024)}]{mediumRecommendationsUsing}
Walsh, M. 2024.
\newblock Recommendations for Using Social Media Data in Research—Whether You’re in English or Info Science.
\newblock Accessed 2024-06-09-2024.

\bibitem[{Wijaya and Yeniterzi(2011)}]{wijaya2011understanding}
Wijaya, D.~T.; and Yeniterzi, R. 2011.
\newblock Understanding semantic change of words over centuries.
\newblock In \emph{Proceedings of the 2011 international workshop on DETecting and Exploiting Cultural diversiTy on the social web}, 35--40. ACM.

\bibitem[{Wilkins, Livingstone, and Levine(2019)}]{wilkins2019whose}
Wilkins, D.~J.; Livingstone, A.~G.; and Levine, M. 2019.
\newblock Whose tweets? The rhetorical functions of social media use in developing the Black Lives Matter movement.
\newblock \emph{British Journal of Social Psychology}, 58(4): 786--805.

\bibitem[{Williams(1985)}]{williams1995}
Williams, R. 1985.
\newblock \emph{"Keywords: A Vocabulary of Culture and Society "}.
\newblock Oxford University Press.

\bibitem[{Xue(2020)}]{malalaBlackLivesMatterYoung}
Xue, H. 2020.
\newblock ""\#BlackLivesMatter: The young Black activists using social media to lead the fight for equality".
\newblock \url{https://assembly.malala.org/stories/young-black-activists-to-follow-on-social}.
\newblock Accessed: 2024-15-05.

\bibitem[{Yan, Chiang, and Lin(2024)}]{Yan_Chiang_Lin_2024}
Yan, M.; Chiang, A.~Y.; and Lin, Y.-R. 2024.
\newblock From Posts to Pavement, or Vice Versa? The Dynamic Interplay between Online Activism and Offline Confrontations.
\newblock \emph{Proceedings of the International AAAI Conference on Web and Social Media}, 18(1): 1687--1701.

\bibitem[{Yourish et~al.(2025)Yourish, Daniel, Datar, White, and Gamio}]{yourish2025}
Yourish, K.; Daniel, A.; Datar, S.; White, I.; and Gamio, L. 2025.
\newblock These Words Are Disappearing in the New Trump Administration.
\newblock \emph{The New York Times}.
\newblock Accessed: 2025-03-07.

\end{thebibliography}

\section{Paper Checklist}

\begin{enumerate}

\item For most authors...
\begin{enumerate}
   \item Would answering this research question advance science without violating social contracts, such as violating privacy norms, perpetuating unfair profiling, exacerbating the socio-economic divide, or implying disrespect to societies or cultures? \answerYes{Yes, and in fact this research seeks to directly intervene into and help to remedy existing socio-economic divides.}
   \item Do your main claims in the abstract and introduction accurately reflect the paper's contributions and scope?\answerYes{Yes, see the Abstract and Introduction.}
   \item Do you clarify how the proposed methodological approach is appropriate for the claims made \answerYes{Yes, see the Background and Prior Work section, along with the various methods sections.}
   \item Do you clarify what are possible artifacts in the data used, given population-specific distributions? \answerYes{Yes, see the Data and Implications and Limitations sections.}
  \item Did you describe the limitations of your work?
\answerYes{Yes, see the Implications and Limitations sections.}
  \item Did you discuss any potential negative societal impacts of your work?
\answerYes{Yes, see the Ethics and Privacy section, which discusses harms potentially brought about by social media research and our steps to mitigate them.}
      \item Did you discuss any potential misuse of your work?
\answerNo{No, because we do not see any potential misuse of our work.}
    \item Did you describe steps taken to prevent or mitigate potential negative outcomes of the research, such as data and model documentation, data anonymization, responsible release, access control, and the reproducibility of findings?
\answerYes{Yes, see the Ethics and Privacy section, which discusses harms potentially brought about by identifying individual users and our steps to mitigate them. Should the paper be accepted, we will provide only derivative data so as to preserve anonymity. We will make our own code publicly available via GitHub.}
  \item Have you read the ethics review guidelines and ensured that your paper conforms to them?
\answerYes{Yes,and we take ethical considerations very seriously.}
\end{enumerate}

\item Additionally, if your study involves hypotheses testing...
\begin{enumerate}
  \item Did you clearly state the assumptions underlying all theoretical results?
\answerNA{NA}
  \item Have you provided justifications for all theoretical results?
\answerNA{NA}
  \item Did you discuss competing hypotheses or theories that might challenge or complement your theoretical results?
\answerNA{NA}
  \item Have you considered alternative mechanisms or explanations that might account for the same outcomes observed in your study?
\answerNA{NA}
  \item Did you address potential biases or limitations in your theoretical framework?
\answerNA{NA}
  \item Have you related your theoretical results to the existing literature in social science?
\answerNA{NA}
  \item Did you discuss the implications of your theoretical results for policy, practice, or further research in the social science domain?
\answerNA{NA}
\end{enumerate}

\item Additionally, if you are including theoretical proofs...
\begin{enumerate}
  \item Did you state the full set of assumptions of all theoretical results?
\answerNA{NA}
	\item Did you include complete proofs of all theoretical results?
\answerNA{NA}
\end{enumerate}

\item Additionally, if you ran machine learning experiments...
\begin{enumerate}
  \item Did you include the code, data, and instructions needed to reproduce the main experimental results (either in the supplemental material or as a URL)?
\answerNA{NA}
  \item Did you specify all the training details (e.g., data splits, hyperparameters, how they were chosen)?
\answerNA{NA}
     \item Did you report error bars (e.g., with respect to the random seed after running experiments multiple times)?
\answerNA{NA}
	\item Did you include the total amount of compute and the type of resources used (e.g., type of GPUs, internal cluster, or cloud provider)?
\answerNA{NA}
     \item Do you justify how the proposed evaluation is sufficient and appropriate to the claims made? 
\answerNA{NA}
     \item Do you discuss what is ``the cost`` of misclassification and fault (in)tolerance?
\answerNA{NA}
  
\end{enumerate}

\item Additionally, if you are using existing assets (e.g., code, data, models) or curating/releasing new assets, \textbf{without compromising anonymity}...
\begin{enumerate}
  \item If your work uses existing assets, did you cite the creators?
\answerYes{Yes, see the Background and Prior Work and Data sections.}
  \item Did you mention the license of the assets?
\answerNA{NA}
  \item Did you include any new assets in the supplemental material or as a URL?
\answerNo{No, but we will put our code on GitHub if the paper is accepted.}
  \item Did you discuss whether and how consent was obtained from people whose data you're using/curating?
\answerYes{Yes, see the Ethics and Privacy section, which discusses our decision to paraphrase rather than obtain consent from individual users; and our use of a ``reasonably public'' threshold for naming specific users, both employe due to concerns over exposure and harm.}
  \item Did you discuss whether the data you are using/curating contains personally identifiable information or offensive content?
\answerYes{Yes, see the Ethics and Privacy section and our response to 5d just above.}
  \item If you are curating or releasing new datasets, did you discuss how you intend to make your datasets FAIR (see \cite{fair})?
\answerNA{NA}
\item If you are curating or releasing new datasets, did you create a Datasheet for the Dataset (see \cite{gebru2021datasheets})? 
\answerNA{NA}
\end{enumerate}

\item Additionally, if you used crowdsourcing or conducted research with human subjects, \textbf{without compromising anonymity}...
\begin{enumerate}
  \item Did you include the full text of instructions given to participants and screenshots?
\answerNA{NA}
  \item Did you describe any potential participant risks, with mentions of Institutional Review Board (IRB) approvals?
\answerNA{NA}
  \item Did you include the estimated hourly wage paid to participants and the total amount spent on participant compensation?
\answerNA{NA}
   \item Did you discuss how data is stored, shared, and deidentified?
\answerNA{NA}
\end{enumerate}

\end{enumerate}

\section{Appendix}

  \begin{table}[h]
      \centering
      \small
      \begin{tabular}{ll}
          \toprule
          Maxima & Additional Candidate Maxima \\
          \midrule
          Sep 02, 2014 & n/a \\
          Dec 25, 2014 & Dec 18, 2014 \\
          Aug 31, 2015 & Mar 14, 2015; Apr 10, 2015; May 04, 2015 \\
          Jul 08, 2016 & n/a \\
          May 27, 2020 & n/a \\
          Aug 07, 2020 & Sep 05, 2020; Sep 25, 2020 \\
          \bottomrule
      \end{tabular}
      \caption{\textbf{Maxima}: Additional candidate maxima that were merged into the final time bin segments.}
      \label{tab:maxima}
  \end{table}
  
  \begin{table}[h]
      \centering
      \small
      \begin{tabular}{ll}
          \toprule
          Community & Short Label \\
          \midrule
          Black Twitter & btwitter \\
          Allies/Academics & allies \\
          BLM Activists & activists \\
          Progressives & progressives \\
          Conservatives & conserv \\
          Black Celebrities & bcelebs \\
          Mixed Celebrities & mcelebs \\
          Center/Left Politicians & centleftpols \\
          Center/Left News Media & centleftnews \\
          Local News & localnews \\
          International Users & intl \\
          \bottomrule
      \end{tabular}
      \caption{\textbf{Labeling Scheme}: Table translating between the long and short labels used throughout this paper.}
      \label{tab:sankey_key}
  \end{table}

\begin{figure*}[h!] 
  \centering
  \includegraphics[width=\linewidth]{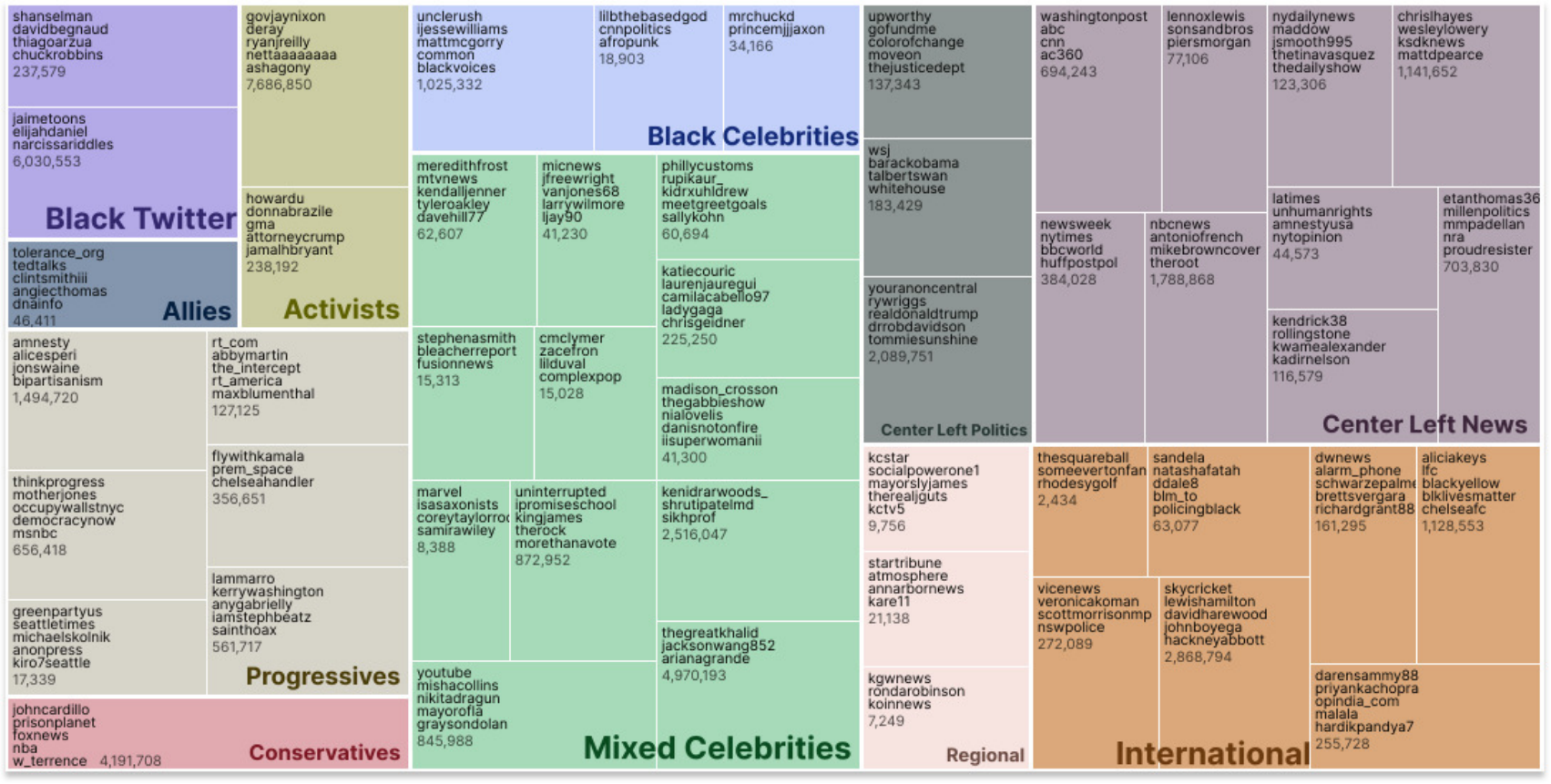}
  \label{fig:treemap}
  \caption{\textbf{Top Community Members}: The most central users of each subcommunity cluster, with color indicating final community grouping. Users with fewer than 3000 followers (as of September 2024) or with inactive accounts are not shown. Note that ranking by in degree leads to certain users (e.g. @realdonaldtrump) being included in communities in which they engage adversarially.}
 \end{figure*}

 \begin{figure}[h]
  \centering
  \includegraphics[width=0.7\linewidth]{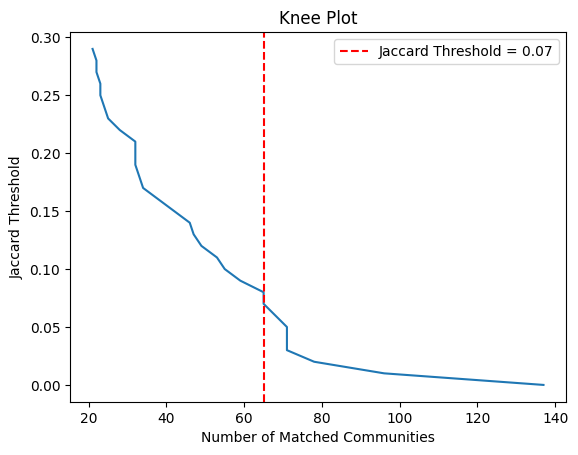}
  \caption{\textbf{Viable thresholds}: A knee plot showing the number of matched communities at Jaccard similarity thresholds between 0 and .3 in increments of .0125.}
  \label{fig:knee}
\end{figure}
\end{document}